# An LSTM-based Plagiarism Detection via Attention Mechanism and a Population-based Approach for Pre-Training Parameters with imbalanced Classes

Seyed Vahid Moravvej[1], Seyed Jalaleddin Mousavirad[2], Mahshid Helali Moghadam[3,4], and Mehrdad Saadatmand[3]

[1]Department of Computer Engineering, Isfahan University of Technology, Isfahan, Iran
[2]Department of Computer Engineering, Hakim Sabzevari Univesity, Sabzevar, Iran
[3]RISE Research Institutes of Sweden, Sweden
[4]Mälardalen University, Västerás, Sweden

**Abstract**. Plagiarism is one of the leading problems in academic and industrial environments, which its goal is to find the similar items in a typical document or source code. This paper proposes an architecture based on a Long Short-Term Memory (LSTM) and attention mechanism called LSTM-AM-ABC boosted by a population-based approach for parameter initialization. Gradient-based optimization algorithms such as back-propagation (BP) are widely used in the literature for learning process in LSTM, attention mechanism, and feed-forward neural network, while they suffer from some problems such as getting stuck in local optima. To tackle this problem, population-based metaheuristic (PBMH) algorithms can be used. To this end, this paper employs a PBMH algorithm, artificial bee colony (ABC), to moderate the problem. Our proposed algorithm can find the initial values for model learning in all LSTM, attention mechanism, and feed-forward neural network, simultaneously. In other words, ABC algorithm finds a promising point for starting BP algorithm. For evaluation, we compare our proposed algorithm with both conventional and population-based methods. The results clearly show that the proposed method can provide competitive performance.

## 1 Introduction

Plagiarism is one of the most important problems in educational institutions such as universities and scientific centers. The purpose of an automated plagiarism detection system is to find similar items at the level of a word, sentence, or document. There are different goals for plagiarism detection. For example, some of these studies only identify duplicate documents [1]. However, low accuracy is a main problem because they do not recognize copied sentences. Several other detectors are designed to find similar source codes in programming environments. It is worthwhile to mention that most research takes into account plagiarism detection at the sentence level [2].

Generally speaking, the methods presented for plagiarism detection are based on statistical methods or in-depth learning. Statistical methods usually utilize Euclidean distance or cosine similarity to calculate the similarity between two items [3-5]. Convolutional Neural Network (CNN) and Recurrent Neural Network (RNN), as two leading deep learning models, have attracted much attention of researchers for plagiarism detection. [6] uses a Siamese CNN to analyze the content of words and selects a representation of a word relevance with its neighbors. In [7], the representation of each word is made using Glove (a word embedding method) [8], and then the representation of the sentences is obtained using a recursive neural network. Finally, similar sentences are identified using cosine similarity. [9] employed two attention-based LSTM networks to extract the representation of two sentences. In [10], the similarity between the sentences is considered for the answer selection task. Two approaches are proposed for this purpose. The first method uses two methods of embedding Language Models (ELMo) [11] and the Bidirectional Encoder Representations from transformers (BERT) and combines them with a transformer encoder. In the second approach, the model is tuned using two pre-trained transformer encoder models. In [12], the authors presented a method based on the context-aligned RNN called CARNN. This paper suggests embedding context information of the aligned words in hidden state generation. They showed that this technique could play an effective role in measuring similarity. In addition, from the literature, there are some few papers focused on the attention mechanism for plagiarism detection [13, 14].

One of the most important reasons for the convergence of neural networks is the initial value of the parameters. The gradient-based algorithms such as back-propagation are extensively used for deep learning models. However, these algorithms have some problems such as sensitivity to initial parameters and getting stuck in local optima [15, 16]. Population-based metaheuristic (PBMH) algorithms can be considered as an alternative to these problems. Artificial Bee Colony (ABC) is an efficient PBMH which has achieved many successes in optimizing a diverse range of applications [17].

In this study, a new architecture, LSTM-AM-ABC, based on the attention mechanism for plagiarism detection at the sentence level is proposed. The proposed algorithm benefits from three main steps including pre-processing, word embedding, and model construction. LSTM-AM-ABC employs LSTM and feed-forward networks as the core model, attention-based mechanism for changing the importance of all inputs, and an ABC algorithm for parameter initialization. Here, the main responsibility of the ABC algorithm is to find a promising point to commence the BP algorithm in LSTM, attention mechanism, and feed-forward network. The proposed model learns two pairs of positive and negative inputs. Negative pairs are dissimilar sentences, while positive pairs are similar sentences. In addition, we use several methods to overcome the data imbalance problem. We evaluate our results on three benchmark datasets based on different criteria. The evaluation results show that the proposed model can be superior to the compared models that use the random value for the parameters.

The remainder of this paper is organized as follows. Section 2 explains briefly the background knowledge. Section 3 illustrates the proposed model, while Section 4 evaluates the proposed model. Finally, Section 5 concludes the paper.

## 2      Background

### 2.1    Long Short-Term Memory (LSTM)

Recurrent neural networks (RNNs) are a type of neural network based on sequential datasets. These networks are utilized in many applications, including natural language processing [18-20], due to their recurrent nature. The network uses a hidden layer to transfer information from $t-1$ to $t$. In addition, an output is generated each time $t$. The hidden layer and the output each time $t$ in the RNN network are calculated as

$$h_t = \theta(W_h h_{t-1} + U_h x_t + b_h) \tag{1}$$

$$y_t = \tau(W_y h_t + b_y) \tag{2}$$

where $x_t, h_t, y_t$ are the input vector, the hidden layer, and the output at time $t$, respectively. $W_h, U_h, W_y$ are the weight matrices, and $b_h, b_y$ are the bias values. $\theta, \tau$ represent the activation functions.

The main problem with a RNN network is the vanishing gradient, meaning that the gradient has a tendency toward zero, which drastically prevents the parameters from being updated. In 1997 [21], LSTM networks were first proposed to solve this problem. An LSTM unit includes an input gate, a memory gate, and an output gate that make it easy to learn long dependencies [13]. The update relations of an LSTM unit in step $t$ are as follows [22]

$$i_t = \sigma(W_i x_t + U_i h_{t-1} + b_i) \tag{3}$$

$$f_t = \sigma(W_f x_t + U_f h_{t-1} + b_f) \tag{4}$$

$$c_t = f_t c_{t-1} + i_t \tanh(W_j x_t + U_j h_{t-1} + b_j) \tag{5}$$

$$o_t = \sigma(W_o x_t + U_o h_{t-1} + b_o) \tag{6}$$

$$h_t = o_t \tanh(c_t) \tag{7}$$

where $i$, $f$, $o$, and $c$ are the input gate, forget gate, output gate and, cell input, respectively. $W \in \mathbb{R}^{h \times d}$. $U \in \mathbb{R}^{h \times h}$, $b \in \mathbb{R}^h$ are network parameters that should be learned during the learning process. Note that the input size $x$ and hidden size $h$ are $d$ and $h$, respectively.

LSTM networks process input from start to finish or vice versa. It has been proven that it can be more effective if the processing is done from both sides simultaneously [23]. Bidirectional Long Short-Term Memory (BLSTM) networks are a type of LSTM networks that process input from both sides and produce two hidden vectors $\vec{h}_t$ and $\overleftarrow{h}_t$. In BLSTM, the combination of two hidden vectors, $h_t = [\vec{h}_t . \overleftarrow{h}_t]$, is considered as the final hidden vector.

Although LSTM networks consider long sequences, they give the same importance to all inputs. It can confuse the network in making decisions. Consider the following sentence: "Despite being from Uttar Pradesh, as she was brought up in Bengal, she is convenient in Bengali". Some words such as "Bengali", "brought up" and "Bengal" should have more weight because it has more related to the word "Bengali". The attention mechanism for this problem was later introduced [24]. In the attention mechanism, for each hidden vector, a coefficient is considered that the final hidden vector is calculated as

$$h = \sum_{t=1}^{T} \alpha_t h_t \tag{8}$$

where $\alpha_t$, $h_t$ is the coefficient of significance and the hidden vector extracted in step $t$. $T$ is the number of inputs.

### 2.2 Artificial bee colony algorithm (ABC)

The artificial bee colony algorithm (ABC) is an effective PBMH algorithm based on collective intelligence of the bee colonies [25]. This algorithm has four main steps, which are described below:

**Initial Population** The initial population of food sources is created with size $N$, and filled randomly. Each $D$-dimensional solution is generated as

$$x_i^j = x_{min}^j + rand(0.1)(x_{max}^j - x_{min}^j) \tag{9}$$

where $i = 1.2.....N$, $j = 1.2.....D$. $x_{min}^j$ min and $x_{max}^j$ are the lower and upper bounds for the dimension $j$, respectively. After initialization, the population is searched for employed bees, spectator bees, and scout bees.

**The Search of the employed bees** In this step, the new solution $v$ is calculated based on the previous solution $x$ as

$$v_i^j = x_i^j + \varphi_i^j (x_i^j - x_k^j) \tag{10}$$

where $j \in \{1.2....D\}$ and $k \in \{1.2.....N\}$ and $k \neq i$ are random indexes. $\varphi_i^j \in [-1.1]$ is a random number. Eq. 10 shows that the new solution $v_i$ is generated by changing a vector element $x_i$. After calculating $v_i$, fitness $v_i$ is calculated. A greedy selection is applied so that if the fitness value of $v_i$ is better, $v_i$ replaces $x_i$. Otherwise, $x_i$ is retained.

**Selection of the onlooker Bees** In this phase, each onlooker bee selects a food source according to the fitness value. The probability of choosing any candidate solution depending on the fitness value is computed as

$$p_i = \frac{fit(x_i)}{\sum_{n=1}^{N} fit(x_n)} \tag{11}$$

where $fit(x_i)$ is the fitness value of solution $x_i$. It is clear that by increasing the value of $fit(x_i)$, the onlooker bee has more probability to choose this food source. After selecting the food source, the onlooker bee will move towards it and produce a new food source in its neighborhood by using Eq. 10.

**Scout bee step** If the position of a food source cannot be further improved than the number of pre-determined cycles ($limit$), a new solution will replace it. Scout bees can discover richer solutions as

$$x_i = x_{min} + rand(0.1)(x_{max} - x_{min}) \tag{12}$$

where $x_{min}$ and $x_{max}$ are the lower and upper bounds of solation $x_i$.

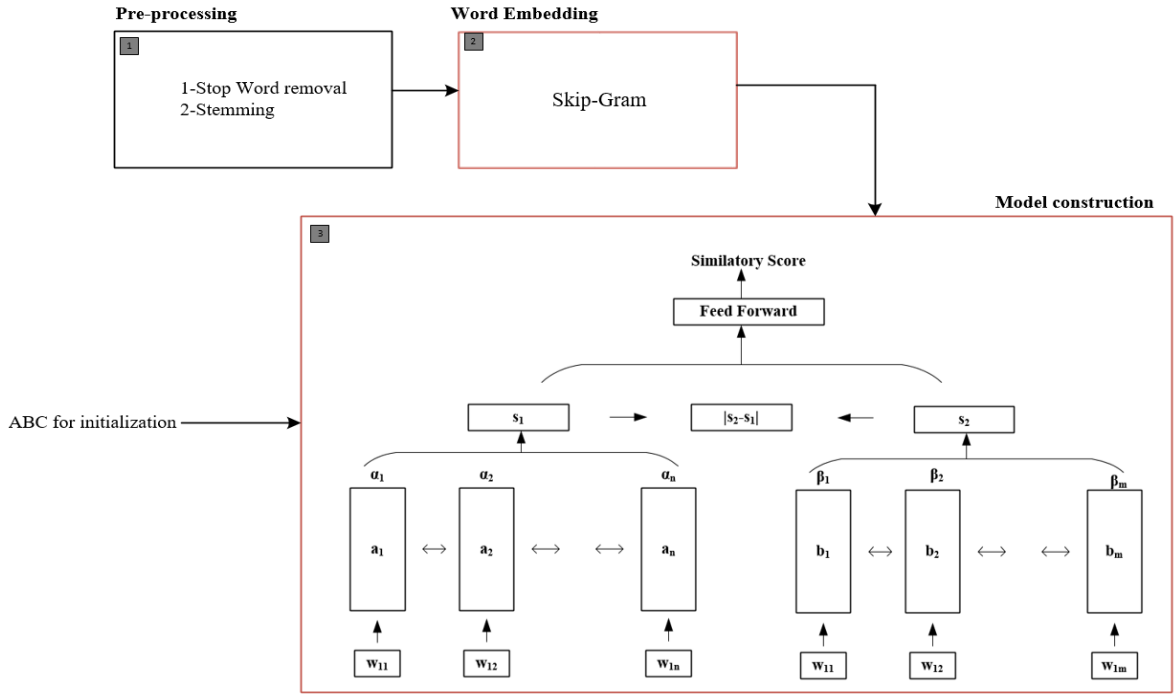

**Fig. 1.** Steps of the proposed model.

## 3    LSTM-AM-ABC approach

The proposed method, LSTM-AM-ABC, consists of three main steps, including pre-processing, word embedding, and model construction (According to Fig. 1). The details of each step are described below.

### 3.1    Pre-processing

Pre-processing means removing unnecessary and unimportant words, which reduces the computational load and increases the speed. Two techniques are used for this purpose.

**Stop-word removal** Words such as 'or' and 'with' lack semantic information due to their repetition and presence in most documents. Eliminating these words plays a crucial role in the performance of plagiarism detection.

**Stemming** The process of returning words to their root form is called the stemming operation (for example, the root form of *looking* is the word of *look*).

### 3.2 Word Embedding

One of the most important steps in natural language processing is word embedding because it is used as input and the embedding of sentences is made based on it. For this purpose, we use the well-known algorithm Skip-Gram [26]. It applies a simple neural network model to learn word vectors. Vectors are carefully generated so that the similarity of the two words can be estimated using a similarity function.

### 3.3 Model construction

This paper proposes a plagiarism detection method, LSTM-AM-ABC, based on LSTM and feed-forward neural network, as the core models, the attention mechanism for altering the importance of the inputs, and ABC for parameter initialization. An LSTM is provided for each sentence $S$. In this research, two pairs of data have been used to learn the model. In positive pair $(S_1, S_2)$, $S_1$ and $S_2$ are two copy sentences. The degree of similarity of the sentences depends on the dataset. In Negative pairs $(S_1, \acute{S}_2)$, $S_1$ and $\acute{S}_2$ are not similar. For a positive pair, class label is one, while for a negative pair, the class label is zero. Let $S_1 = \{w_{11}. w_{12}. \ldots. w_{1n}\}$ and $S_2 = \{w_{21}. w_{22}. \ldots. w_{2m}\}$ be two sentences, where $w_{ij}$ is the $j-th$ word in $i-th$ sentence. The two sentences $S_1$ and $S_2$ are limited to $n$ and $m$ words, respectively. The embedding of sentences $s_1$ and $s_2$ is formulated based on the attention mechanism as

$$s_1 = \sum_{i=1}^{n} \alpha_i\, h_{a_i} \tag{13}$$

$$s_2 = \sum_{i=1}^{m} \beta_i\, h_{b_i} \tag{14}$$

where $h_{a_i} = [\vec{h}_{a_i}. \overleftarrow{h}_{a_i}] \in \mathbb{R}^{2d_1}$, $h_{b_i} = [\vec{h}_{b_i}. \overleftarrow{h}_{b_i}] \in \mathbb{R}^{2d_2}$ are the $i-th$ output in BLSTM. Each BLSTM output plays a role in the output with a coefficient in the range [0,1]. These coefficients are calculated for both networks as

$$\alpha_i = \frac{e^{u_i}}{\sum_{i=1}^{n} e^{u_i}} \tag{15}$$

$$\beta_i = \frac{e^{v_i}}{\sum_{i=1}^{m} e^{v_i}} \tag{16}$$

$$u_i = tanh(W_u h_{a_i} + b_u) \tag{17}$$

$$v_i = tanh(W_v h_{b_i} + b_u) \tag{18}$$

where $W_u \in \mathbb{R}^{2d_1}. b_v \in \mathbb{R}$, $W_v \in \mathbb{R}^{2d_2}$. and $b_v \in \mathbb{R}$ are the parameters of the attention mechanism for two sentences. After calculating the embedding of sentences, they, along with their differences $|s_2 - s_1|$, enter a feed-forward network, and their similarity is calculated.

### 3.3.1 Parameter optimization

There is a plethora of parameters in the proposed model including parameters in LSTM, feed-forward networks, and attention mechanism. This paper proposes a novel approach for parameter initialization using ABC algorithm. To this end, two main issues should be considered including encoding strategy and fitness function. Encoding strategy represents the structure of each candidate solution, while fitness function is responsible to calculate the quality of each candidate solution.

#### 3.3.1.1 Encoding strategy

The proposed model consists of three main parts including two LSTM networks, two attention mechanism systems, and one feed-forward network. Fig.2 shows a typical encoding strategy for a two-layer feed-forward network, two single-layer LSTM networks and two attention mechanisms.

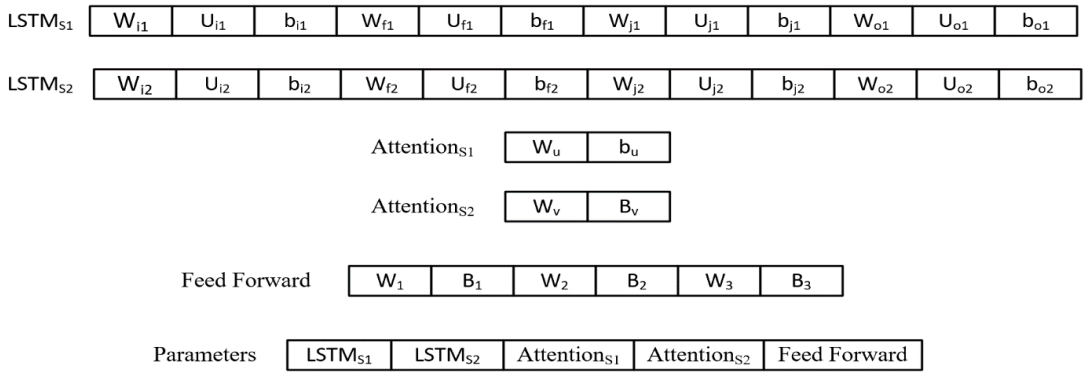

**Fig. 2.** Illustration of encoding strategy.

#### 3.3.1.2 Fitness function

Fitness function calculates the quality of each candidate solution. In this paper, we propose an objective function based on similarity as

$$Fitness = \frac{1}{1+\sum_{i=0}^{N}(y_i-\tilde{y}_i)^2} \qquad (19)$$

where $N$ is training data, $y_i$ is the $i-th$ target, and $\tilde{y}_i$ is the predicted similarity value for the $i-th$ data. The goal of optimization here is to find the optimal initial seeds for the BP algorithm.

### 3.3.2 Imbalance Classification

One of the most important challenges of machine learning is the problem of data imbalance. Data imbalance means that the number of data in the classes is not the same and, one of the classes (or even more) has more data. It reduces system performance because it causes the classifier to bias the output to one side. In our case, the problem of imbalance is due to a large number of negative pairs. This paper proposes a combination of imbalance methods to tackle the problem.

**Augmentation** The goal is to increase the positive pair. For this purpose, we combine the embedding of each word in the sentence with Gaussian noise and produce new sentences that are similar to positive sentences.

**Penalty** In this technique, the minority and majority class error values are applied with different coefficients in the Loss function. Eq. 20 illustrates the concept of this technique

$$Loss = \alpha Loss_{C_1} + \beta Loss_{C_2} \qquad (20)$$

where $C_1$ and $C_2$ are the minority and majority classes, respectively, and the coefficients $\alpha$ and $\beta$ are the importance of the Loss function. If $\alpha$ and $\beta$ are equal, we have a common classification problem.

## 4 Experiments and Analysis

In this section, we evaluate LSTM-AM-ABC algorithm compared to other competitors and different criteria.

### 4.1 Corpus

Plagiarism detection is a classification problem

$$sim(S_1.S_2) = \begin{cases} \geq \varepsilon & S_1 \text{ is a copy of } S_2 \\ < \varepsilon & S_1 \text{ is not a copy of } S_2 \end{cases} \qquad (21)$$

According to Eq. 21, when the proposed model detects the degree of similarity of $S_1$ and $S_2$ above 0.5, a copy is detected. In this research, three common plagiarism datasets are utilized for this purpose.

**SemEval2014** This dataset is taken from the Sentences Involving Com-positional Knowledge (SICK) dataset [27] for semantic evaluation of English sentences. It has 5000 pairs of sentences for training, and 5,000 sentences for testing. Each pair of sentences has a similarity label ∈ {1.2.3.4.5}, which 1 means that the sentences are irrelevant and 5 meaning the most similar sentences. We consider the label = 0 as class 0 and the label ∈ {2.3.4.5} as class 1.

**STS Semantic** Text Similarity (STS) [9] is based on image captions, news headlines, and user forums. This database has 6928 sentences for training, and 1500 sentences for testing. To make unrelated pairs, we put unrelated sentences together.

**MSRP** Microsoft Research Paraphrase corpus (MSRP) [13] is a paraphrasing corpus containing 4076 pairs of sentences for training and 1725 pairs of sentences for testing. The sentences in this dataset are tagged by humans.

### 4.2 Metrics

**Recall** For plagiarism systems, it is vital not to recognize sentences that are copies so that copies can easily pass through the filter without being detected. The recall criterion is one of the valuable criteria for this problem because this criterion considers the number of copied sentences is not recognized. The recall metric is defined as

$$Recall = \frac{TP}{TP+FN} \qquad (22)$$

where $TP$ is the number of states that the system has correctly detected the copy, and $FN$ is the number of states that the system has not detected the actual copy.

Pearson's correlation This criterion distinguishes negative correlation and negative correlation numerically in the interval [-1,1] as

$$r = \frac{Cov(sim_{y'}, sim_y)}{\sqrt{var(sim_{y'})var(sim_y)}} \tag{23}$$

According to this relation, the value of r in the range [0,0.2], [0.2,0.4] and [0.4,0.6] means zero correlation, weak correlation and moderated correlation, respectively [13].

**Mean Square Error(MSE)** The MSE criterion indicates the difference between the degree of actual and predicted similarity defined as

$$\text{MSE} = sum((sim_{y'} - sim_y)^2) \tag{24}$$

### 4.3  Results

We compare the proposed method with a series of previous researches. To this end, we use k-fold cross-validation (k=10 or 10CV) in all experiments, in which the dataset is divided into k subsets. One of the subsets is used for test data, while the remaining is employed for training. This procedure is repeated k times, and all data is used exactly once for testing. We report statistical results including mean, standard deviation, and median for each criterion and each dataset. Table 1 shows the parameter settings of the proposed model. In addition, ε was set to 0.515, 0.525, and 0.52 for SemEval2014, STS dataset, and MSRP datasets, respectively.

We compare our method with seven methods, including Siamese CNN+LSTM [6], Recurrent NN, 100D, dropout 20 [7], CETE [10], CA-RNN [12], STS-AM [9], AttSiaBiLSTM [13], and LSTM + FNN + attention [14]. Also, we compare our algorithm with the LSTM-AM that employ random number as the initial point of parameters to show that our proposed ABC approach can effectively improve the results. The results of the evaluation are shown in Tables 2, 3, and 4. For the SemEval2014 dataset, LSTM-AM-ABC has been able to overcome other methods in the recall and r criteria. Comparing LSTM-AM-ABC with LSTM-AM clearly indicates the effectiveness of our proposed initialization approach. For the STS dataset, LSTM-AM-ABC again presented the best results compared to other algorithms. By comparing LSTM-AM-ABC, we can observe that LSTM-AM-ABC could decrease the error more than 40%, indicating the effectiveness of initialization approach. The results of Table 4 are consistent with other tables. For MSRP dataset, our proposed algorithm obtained the highest mean recall followed by CETE algorithm.

**Table 1.** Parameter setting for the proposed model.

| Parameter | Value |
| --- | --- |
| max sentence length | 100 |
| embedding dim | 80 |
| penalty rate for positive pairs (α) | 1 |
| penalty rate for negative pairs (β) | 0.5 |
| blstm hidden dim | 50 |
| dense hidden layer | 3 |
| dense hidden dim | 256,128,64 |

**Table 2.** 10CV classification results on SemEval2014 dataset.

| Method | Recall | | | MSE | | | r | | |
|---|---|---|---|---|---|---|---|---|---|
| | mean | std.dev. | median | mean | std.dev. | median | mean | std.dev. | median |
| Siamese CNN+LSTM [6] | 82.136 | 5.360 | 86.403 | 0.286 | 0.096 | 0.292 | 0.506 | 0.314 | 0.741 |
| Recurrent NN, 100D, dropout 20 [7] | 82.563 | 2.068 | 83.128 | 0.164 | 0.092 | 0.183 | 0.721 | 0.217 | 0.863 |
| CETE [10] | 91.153 | 1.143 | 92.119 | 0.055 | 0.063 | 0.057 | 0.709 | 0.206 | 0.817 |
| CA-RNN [12] | 85.436 | 1.890 | 85.763 | 0.034 | 0.067 | 0.062 | 0.754 | 0.145 | 0.786 |
| STS-AM [9] | 88.477 | 1.683 | 89.809 | 0.059 | 0.041 | 0.060 | 0.791 | 0.262 | 0.826 |
| AttSiaBiLSTM [13] | 84.016 | 0.935 | 84.639 | 0.036 | 0.053 | 0.039 | 0.759 | 0.249 | 0.790 |
| LSTM + FNN + attention [14] | 86.103 | 2.360 | 88.509 | 0.062 | 0.092 | 0.099 | 0.776 | 0.174 | 0.820 |
| LSTM-AM | 93.129 | 3.390 | 94.208 | 0.076 | 0.068 | 0.089 | 0.812 | 0.170 | 0.912 |
| LSTM-AM-ABC | 95.268 | 1.791 | 97.018 | 0.053 | 0.047 | 0.062 | 0.804 | 0.183 | 0.963 |

**Table 3.** 10CV classification results on STS dataset.

| Method | Recall | | | MSE | | | r | | |
|---|---|---|---|---|---|---|---|---|---|
| | mean | std.dev. | median | mean | std.dev. | median | mean | std.dev. | median |
| Siamese CNN+LSTM [6] | 85.153 | 4.712 | 88.106 | 0.125 | 0.088 | 0.147 | 0.549 | 0.229 | 0.570 |
| Recurrent NN, 100D, dropout 20 [7] | 86.100 | 3.190 | 88.247 | 0.099 | 0.081 | 0.121 | 0.769 | 0.187 | 0.775 |
| CETE [10] | 95.014 | 1.371 | 96.053 | 0.043 | 0.082 | 0.056 | 0.784 | 0.225 | 0.809 |
| CA-RNN [12] | 89.056 | 0.441 | 89.610 | 0.029 | 0.054 | 0.035 | 0.771 | 0.215 | 0.797 |
| STS-AM [9] | 92.101 | 3.943 | 94.105 | 0.048 | 0.072 | 0.059 | 0.820 | 0.042 | 0.826 |
| AttSiaBiLSTM [13] | 86.283 | 1.800 | 87.120 | 0.031 | 0.062 | 0.039 | 0.782 | 0.162 | 0.798 |
| LSTM + FNN + attention [14] | 89.156 | 1.089 | 89.664 | 0.049 | 0.070 | 0.054 | 0.801 | 0.140 | 0.819 |
| LSTM-AM | 96.206 | 3.610 | 97.421 | 0.061 | 0.052 | 0.072 | 0.832 | 0.129 | 0.840 |
| LSTM-AM-ABC | 97.410 | 3.811 | 98.163 | 0.041 | 0.051 | 0.054 | 0.840 | 0.091 | 0.849 |

The proposed algorithm employs ABC in conjunction with BP algorithm for training. In the following, we indicate the proposed training algorithm is effective compared to others. To have a fair comparison, we fix all remaining parts of our proposed algorithm including LSTM, feedforward network, attention-based mechanism and only the trainer is changed. To this end, we compare our proposed trainer with five conventional algorithms, including Gradient Descent with simple Momentum (GDM) [28], Gradient Descent with Adaptive learning rate backpropagation (GDA) [29], Gradient Descent with Momentum and Adaptive learning rate backpropagation (GDMA) [30], One-Step Secant backpropagation (OSS) [31], and Bayesian Regularization backpropagation (BR) [32], And four metaheuristic algorithms, including Grey Wolf Optimization (GWO) [33], Bat Algorithm (BA) [34], Cuckoo Optimization Algorithm (COA) [35], and Whale Optimization Algorithm (WOA) [36]. For all metaheuristic algorithms, the population size and number of function evaluations are set to 50 and 20,000, respectively. Other parameter settings can be seen in Table. 5.

**Table 4.** 10CV classification results on MSRP dataset.

| Method | Recall | | | MSE | | | r | | |
|---|---|---|---|---|---|---|---|---|---|
| | mean | std.dev. | median | mean | std.dev. | median | mean | std.dev. | median |
| Siamese CNN+LSTM [6] | 87.089 | 2.790 | 88.119 | 0.096 | 0.093 | 0.106 | 0.713 | 0.181 | 0.723 |
| Recurrent NN, 100D, dropout 20 [7] | 89.207 | 1.341 | 89.690 | 0.043 | 0.080 | 0.051 | 0.787 | 0.210 | 0.817 |
| CETE [10] | 97.296 | 0.910 | 97.429 | 0.016 | 0.073 | 0.024 | 0.819 | 0.189 | 0.820 |
| CA-RNN [12] | 90.179 | 1.207 | 91.092 | 0.018 | 0.091 | 0.043 | 0.787 | 0.163 | 0.799 |
| STS-AM [9] | 91.396 | 1.179 | 91.647 | 0.057 | 0.087 | 0.069 | 0.809 | 0.187 | 0.816 |
| AttSiaBiLSTM [13] | 87.493 | 3.410 | 89.190 | 0.025 | 0.067 | 0.049 | 0.801 | 0.018 | 0.809 |
| LSTM + FNN + attention [14] | 89.269 | 2.107 | 91.018 | 0.029 | 0.058 | 0.057 | 0.818 | 0.196 | 0.839 |
| LSTM-AM + random weight | 95.208 | 1.874 | 96.410 | 0.069 | 0.059 | 0.072 | 0.829 | 0.100 | 0.841 |
| LSTM-AM-ABC | 97.379 | 1.270 | 97.941 | 0.035 | 0.064 | 0.057 | 0.857 | 0.073 | 0.869 |

**Table 5.** Parameter setting for metaheuristic algorithms.

| algorithm | parameter | value |
|---|---|---|
| ABC [37] | maximum number of failures | population size × dimensions of each solution |
| BAT [34] | constant for loudness update | 0.5 |
|  | constant for an emission rate update | 0.5 |
|  | initial pulse emission rate | 0.001 |
| COA [35] | Discovery rate of alien solutions | 0.25 |
| GWO [33] | no parameters |  |
| WOA [36] | b | 1 |

**Table 6.** Results of 10CV classification of metaheuristic algorithms on SemEval2014 dataset.

| Algorithm | Recall | | | MSE | | | r | | |
|---|---|---|---|---|---|---|---|---|---|
|  | mean | std.dev. | median | mean | std.dev. | median | mean | std.dev. | median |
| LSTM-AM-GDM | 89.126 | 1.142 | 90.250 | 0.072 | 0.096 | 0.084 | 0.774 | 0.125 | 0.850 |
| LSTM-AM-GDA | 88.473 | 1.480 | 88.892 | 0.076 | 0.024 | 0.081 | 0.759 | 0.107 | 0.800 |
| LSTM-AM-GDMA | 88.421 | 3.189 | 90.547 | 0.082 | 0.035 | 0.086 | 0.752 | 0.114 | 0.792 |
| LSTM-AM-OSS | 87.634 | 5.103 | 89.420 | 0.085 | 0.042 | 0.089 | 0.743 | 0.120 | 0.810 |
| LSTM-AM-BR | 92.169 | 1.300 | 92.962 | 0.070 | 0.029 | 0.075 | 0.788 | 0.103 | 0.824 |
| LSTM-AM-GWO | 90.145 | 1.250 | 91.123 | 0.072 | 0.025 | 0.077 | 0.775 | 0.102 | 0.836 |
| LSTM-AM-BAT | 91.160 | 0.146 | 91.532 | 0.068 | 0.012 | 0.072 | 0.780 | 0.094 | 0.792 |
| LSTM-AM-COA | 92.790 | 1.365 | 93.475 | 0.057 | 0.062 | 0.061 | 0.792 | 0.106 | 0.821 |
| LSTM-AM-WOA | 91.756 | 1.250 | 92.750 | 0.061 | 0.020 | 0.065 | 0.782 | 0.112 | 0.835 |
| LSTM-AM-ABC | 95.268 | 1.791 | 97.018 | 0.053 | 0.047 | 0.062 | 0.804 | 0.183 | 0.963 |

**Table 7.** Results of 10CV classification of metaheuristic algorithms on STS dataset.

| Algorithm | Recall | | | MSE | | | r | | |
|---|---|---|---|---|---|---|---|---|---|
|  | mean | std.dev. | median | mean | std.dev. | median | mean | std.dev. | median |
| LSTM-AM-GDM | 90.100 | 1.967 | 91.250 | 1.020 | 0.047 | 1.046 | 0.775 | 0.092 | 0.781 |
| LSTM-AM-GDA | 92.580 | 1.500 | 93.473 | 0.075 | 0.042 | 0.082 | 0.791 | 0.096 | 0.802 |
| LSTM-AM-GDMA | 91.140 | 2.450 | 93.485 | 0.062 | 0.051 | 0.067 | 0.782 | 0.112 | 0.791 |
| LSTM-AM-OSS | 89.263 | 3.593 | 91.530 | 0.066 | 0.068 | 0.071 | 0.760 | 0.132 | 0.773 |
| LSTM-AM-BR | 96.256 | 2.850 | 97.253 | 0.064 | 0.062 | 0.070 | 0.836 | 0.115 | 0.842 |
| LSTM-AM-GWO | 93.020 | 1.740 | 93.863 | 0.025 | 0.052 | 0.034 | 0.818 | 0.082 | 0.826 |
| LSTM-AM-BAT | 95.418 | 1.425 | 95.920 | 0.072 | 0.059 | 0.079 | 0.831 | 0.079 | 0.838 |
| LSTM-AM-COA | 96.520 | 3.475 | 97.835 | 0.053 | 0.072 | 0.062 | 0.838 | 0.121 | 0.846 |
| LSTM-AM-WOA | 93.120 | 2.148 | 95.128 | 0.061 | 0.068 | 0.069 | 0.824 | 0.118 | 0.836 |
| LSTM-AM-ABC | 97.410 | 3.811 | 98.163 | 0.051 | 0.047 | 0.054 | 0.840 | 0.091 | 0.849 |

**Table 8.** Results of 10CV classification of metaheuristic algorithms on MSRP dataset.

| Algorithm | Recall | | | MSE | | | r | | |
|---|---|---|---|---|---|---|---|---|---|
|  | mean | std.dev. | median | mean | std.dev. | median | mean | std.dev. | median |
| LSTM-AM-GDM | 89.180 | 2.893 | 90.658 | 0.082 | 0.090 | 0.089 | 0.782 | 0.126 | 0.819 |
| LSTM-AM-GDA | 93.185 | 2.459 | 94.150 | 0.072 | 0.089 | 0.081 | 0.810 | 0.093 | 0.827 |
| LSTM-AM-GDMA | 90.163 | 3.485 | 92.635 | 0.079 | 0.093 | 0.086 | 0.786 | 0.150 | 0.824 |
| LSTM-AM-OSS | 91.280 | 2.963 | 92.895 | 0.064 | 0.082 | 0.072 | 0.792 | 0.092 | 0.080 |
| LSTM-AM-BR | 96.000 | 1.285 | 96.142 | 0.042 | 0.060 | 0.050 | 0.819 | 0.035 | 0.082 |
| LSTM-AM-GWO | 95.183 | 1.590 | 95.740 | 0.053 | 0.072 | 0.062 | 0.804 | 0.020 | 0.816 |
| LSTM-AM-BAT | 96.180 | 2.010 | 97.005 | 0.038 | 0.094 | 0.049 | 0.832 | 0.081 | 0.843 |
| LSTM-AM-COA | 96.138 | 2.583 | 96.935 | 0.045 | 0.085 | 0.052 | 0.826 | 0.070 | 0.831 |
| LSTM-AM-WOA | 92.052 | 1.390 | 93.128 | 0.068 | 0.072 | 0.076 | 0.796 | 0.068 | 0.802 |
| LSTM-AM-ABC | 97.379 | 1.270 | 97.941 | 0.035 | 0.064 | 0.057 | 0.857 | 0.073 | 0.869 |

The results of the proposed algorithms compared to other trainers are shown Tables 6, 7, and 8. For the SemEval2014 dataset, as expected, metaheuristic algorithms generally work better than conventional algorithms. BR algorithm has been able to overcome metaheuristic algorithms including GWO, BAT, COA, and WOA. It can be seen that LSTM-AM-ABC outperformed all metaheuristic and conventional algorithms. In SemEval2014 datasets, the proposed trainer can reduce error more than 34% compared to the second best algorithm, LSTM-AM-BR. Such a difference exists in two other datasets, so that LSTM-AM-ABC improved error more than 25% and 32% for STS and MSRP datasets, respectively.

## 5 Conclusions

The goal of plagiarism is to find the similar items in a typical document or source code. This paper proposes a novel model for plagiarism detection based on LSTM-based architecture, feedforward neural networks, attention mechanism incorporating a population-based approach for parameter initialization (LSTM-AM-ABC). Gradient-based optimization algorithms such as back-propagation (BP) is so popular for learning process in LSTM, attention mechanism, and feed-forward neural network, whereas have some problems such as being sensitive in the initial conditions. Therefore, this paper proposed an artificial bee colony (ABC) mechanism to find initial seed for BP algorithm. ABC algorithm is employed on LSTM, attention mechanism, and feed-forward neural network, simultaneously. For evaluation, we compared LSTM-AM-ABC with both conventional and population-based methods. The experimental results on three datasets demonstrate that LSTM-AM-ABC is superior to previous systems. As future work, we intend to provide a suitable solution for imbalanced classification. A simple solution could be to use reinforcement learning. Reinforcement learning can be effective because of the rewards and punishments involved.